\documentclass[11pt]{article}

\usepackage[final]{acl}
\usepackage{times}
\usepackage{latexsym}
\usepackage[T1]{fontenc}
\usepackage[utf8]{inputenc}
\usepackage{microtype}
\usepackage{inconsolata}
\usepackage{graphicx}
\usepackage{amsmath}
\usepackage{amssymb}
\usepackage{amsthm}
\usepackage{booktabs}
\usepackage{multirow}
\usepackage{makecell}
\usepackage{xcolor}
\usepackage{colortbl}
\usepackage{algorithm}
\usepackage{algpseudocode}
\usepackage{subcaption}
\usepackage{xspace}
\usepackage[most]{tcolorbox}
\usepackage{enumitem}
\usepackage{fontawesome5}

\newcommand{\ourmethod}{\textsc{Mage}\xspace}

\title{Beyond Semantic Organization: Memory as Execution State Management for Long-Horizon Agents}

\author{
 \textbf{Yaoqi Chen\textsuperscript{1,2}},
 \textbf{Haibin Lai\textsuperscript{2}},
 \textbf{Yuru Feng\textsuperscript{2,4}},
 \textbf{Chuyu Han\textsuperscript{3}},
 \textbf{Qianxi Zhang\textsuperscript{2,$\dagger$}},
 \textbf{Baotong Lu\textsuperscript{2}},
 \\
 \textbf{Menghao Li\textsuperscript{2}},
 \textbf{Xinjiang Wang\textsuperscript{2}},
 \textbf{Zewen Jin\textsuperscript{1,$\dagger$}},
 \textbf{Hao Wu\textsuperscript{3}},
 \textbf{Cheng Li\textsuperscript{1}},
 \textbf{Qi Chen\textsuperscript{2}}
\\
 \textsuperscript{1}University of Science and Technology of China,
 \textsuperscript{2}Microsoft,
\\
 \textsuperscript{3}Nanjing University,
 \textsuperscript{4}University of California, San Diego
\\
{\normalsize
 \faIcon[regular]{envelope}\
 \href{mailto:yaoqi_chen@mail.ustc.edu.cn}{\texttt{yaoqi\_chen@mail.ustc.edu.cn}},
 \href{mailto:qianxi.zhang@microsoft.com}{\texttt{qianxi.zhang@microsoft.com}}
}
}

\begin{document}
\maketitle

\begingroup
\renewcommand{\thefootnote}{$\dagger$}
\footnotetext{Corresponding authors.}
\endgroup
\setcounter{footnote}{0}

\begin{abstract}
LLM-based agents increasingly tackle long-horizon tasks with interdependent decisions, where each action reshapes future constraints and intermediate errors can cascade. Existing RAG and agent memory systems organize histories by semantic similarity, retrieving content-relevant entries at decision time. We argue that this design mismatches execution-state dependencies: it fragments decision trajectories and mixes valid and erroneous traces, hindering coherent state reconstruction and error isolation. We propose \ourmethod (\textbf{M}emory as \textbf{A}gent-\textbf{G}uided \textbf{E}xploration), an active execution-state manager that stores interactions in a hierarchical state tree. The agent derives its state from the active root-to-current path, combining subgoal summaries, recent traces, and hints from prior branches. Four coupled operations maintain the tree: \texttt{Grow} records new traces, \texttt{Compress} summarizes completed subgoals, \texttt{Maintain} validates summaries, and \texttt{Revise} restores a target boundary and resumes on a new branch. This design bounds context growth while preserving state integrity and isolating flawed segments from the active path. Experiments on MemoryArena show that \ourmethod improves the average task success rate by $7.8$--$20.4$~pp over baselines, while reducing token consumption by 55.1\%.
Code is available at: \href{https://github.com/Komorebi660/MAGE}{\texttt{https://github.com/Komorebi660/MAGE}}.
\end{abstract}

\section{Introduction}
\label{sec:intro}


With the growing ability of large language models (LLMs) to interact with complex environments through tool use and multi-step reasoning, LLM-based agents are increasingly deployed for long-horizon tasks with interdependent decisions~\citep{webshop2022, travelplanner2024, webarena2024, hierarchical2025, llmbasedmultiagentsystemssoftware2025}. These tasks involve hundreds of steps where each action reshapes future choices, and intermediate errors can cascade to invalidate subsequent progress. Unlike recall-oriented memory benchmarks that answer questions over past conversations or agentic traces~\citep{locomo2024, longmemeval2024, amabench2026}, the interdependent long-horizon agent tasks we study require maintaining a coherent, evolving execution state, as each decision depends on the cumulative outcome of prior steps.

\begin{figure}[t!]
\centering
\includegraphics[width=.9\linewidth]{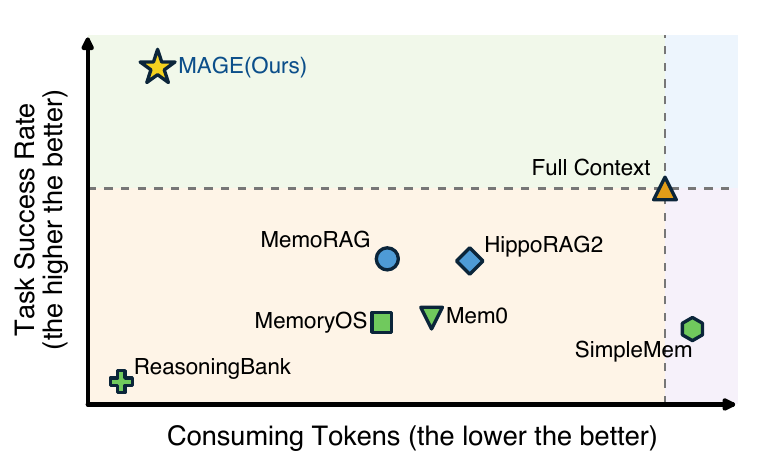}
\caption{\textbf{Paradigm comparison on long-horizon agent tasks} (MemoryArena). Full-context approach achieves strong task success rate but with high token cost, whereas baselines reduce context at the risk of losing state dependencies and underperforming. By managing memory as an execution-state tree, \ourmethod reaches the ideal upper-left region with the highest task success rate and fewer tokens than full-context.}
\label{fig:pr_tokens}
\end{figure}


This requirement becomes harder as exploration history grows beyond the model's effective context window~\citep{memgpt2024, lostinthemiddle2024}. To address this, recent works introduce memory systems~\citep{mem02025, amem2025, zep2025, memoryos2025, hiagent2024, gmemory2025, simplemem2025} that record past information as compact entries and retrieve relevant ones on demand. Yet recent benchmarks reveal a counter-intuitive pattern: these systems often fail to improve long-horizon agent performance and sometimes underperform approaches that simply retain the full history in context~\citep{memoryarena2026, amabench2026}. As shown in Figure~\ref{fig:pr_tokens}, many such systems consume substantial tokens while still trailing the full-context approach.


We argue that a key cause lies in the shared design philosophy. Although these systems vary in their data structures, ranging from flat vector stores to entity-relation graphs to hierarchical architectures, they generally rely on \emph{semantic relationships} to organize and retrieve information, surfacing entries by their content relevance to the current query rather than their role in the execution trajectory.

\begin{figure*}[t]
\centering
\includegraphics[width=\textwidth]{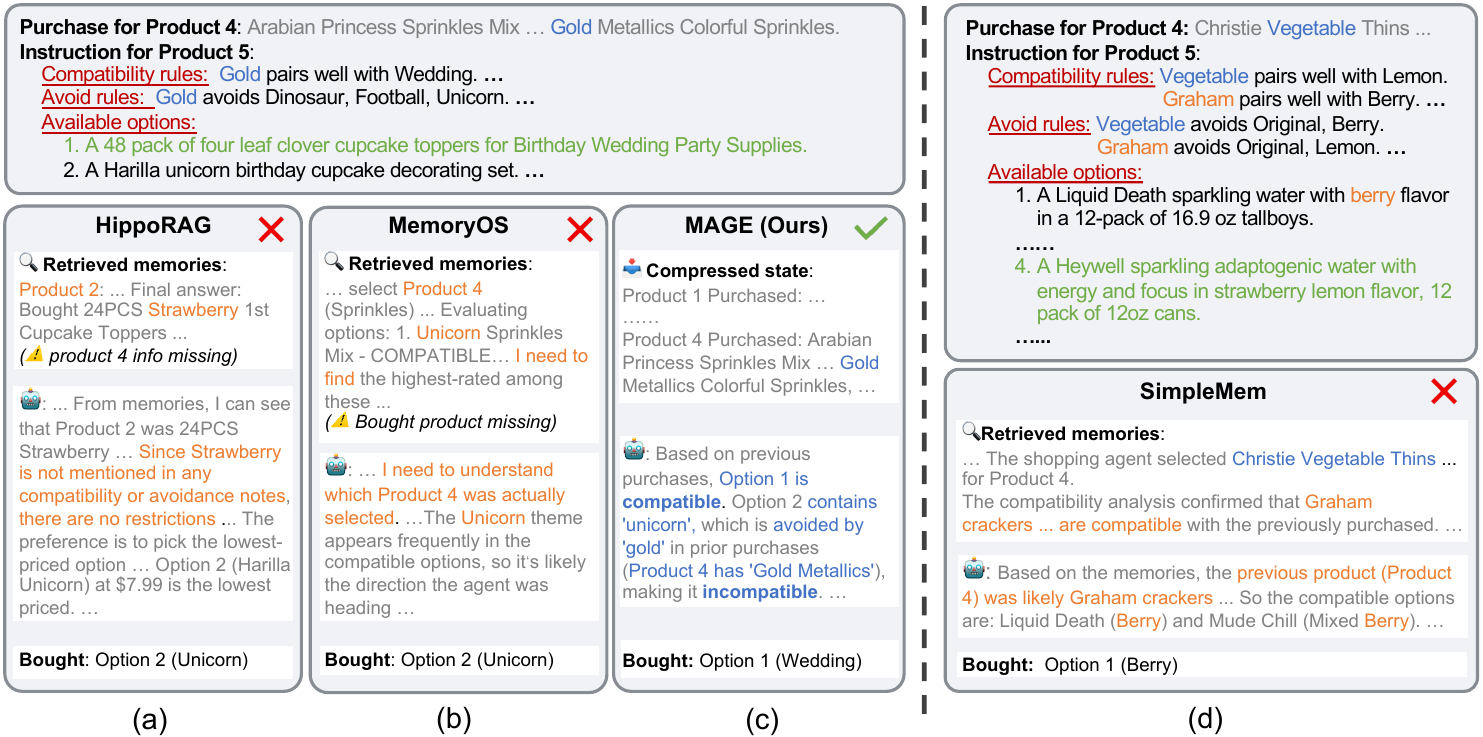}
\caption{\textbf{Case study of baseline failures in MemoryArena shopping tasks}~\citep{memoryarena2026}. In this task, each purchase must satisfy constraints induced by previously bought products. Cases (a)--(c) show state fragmentation: HippoRAG~\citep{hipporag2025} retrieves only Product~2 information, while MemoryOS~\citep{memoryos2025} retrieves Product~4 exploration traces but not the final purchased item; both miss that Product~4 contains \textit{gold} and choose an incompatible \textit{unicorn}-themed option. In contrast, \ourmethod preserves the execution state and selects the compatible \textit{wedding} option. Case (d) shows error contamination, where SimpleMem~\citep{simplemem2025} retrieves mixed evidence from correct and incorrect trajectories and buys a \textit{berry}-flavored item that violates the current \textit{vegetable}-related avoidance rule.}
\label{fig:case_study_baseline_failure}
\end{figure*}

Such similarity-driven organization overlooks the challenge of maintaining a coherent execution state, leading to two recurring problems when handling interdependent long-horizon tasks. First, it causes \textbf{state fragmentation} that weakens execution state integrity. The agent's execution state is built up through a chain of dependent decisions where each step is conditioned on the context established by prior steps. Existing systems, even those with graph structures, organize this state as entries linked by semantic or topical relationships rather than state dependencies, discarding critical execution context that binds them together. As a result, the system may fail to reconstruct a complete, coherent execution state, leading to erroneous actions based on incomplete information (Figure~\ref{fig:case_study_baseline_failure}(a)--(b)).

Second, it hinders \textbf{effective error isolation}. Similarity-based memory mixes entries from different trajectories or exploration attempts in the same relevance space, so erroneous and valid traces can be surfaced together and contaminate subsequent reasoning (Figure~\ref{fig:case_study_baseline_failure}(d)). Without explicit path structure and revision boundaries, it is also difficult to trace an error back to its origin or isolate the affected segment, allowing errors to propagate and accumulate over the course of execution.


These observations suggest that memory for interdependent long-horizon agents should shift from a similarity-driven archive to an \emph{execution-state manager}. To this end, we propose \ourmethod (\textbf{M}emory as \textbf{A}gent-\textbf{G}uided \textbf{E}xploration), which treats memory as an execution state structure rather than a pool of retrievable facts. \ourmethod organizes the agent's history as a persistent two-layer hierarchical state tree. The bottom layer records the step-by-step action-observation trace, while the top layer stores summaries generated at subgoal or decision boundaries. This boundary-aware compression reduces context without interrupting an active trace or breaking execution-state integrity. The current execution state is read from the active tree path instead of being assembled from semantically similar entries, combining compressed state, recent raw state, and execution hints from sibling branches. This path-based representation addresses state fragmentation by keeping the agent-facing state coherent while still bounding the context size.

Building on this tree, \ourmethod further supports error isolation by making memory an agent-manipulable object rather than a shared pool of mixed entries. Through a closed-loop execution cycle, \texttt{Grow} extends the raw trace and \texttt{Compress} summarizes the accumulated trace at subgoal or decision boundaries. Before a new summary becomes trusted memory, \texttt{Maintain} validates the summary and its underlying trace against the task, catching missing information or execution errors before they propagate. If an error is detected, \texttt{Revise} restores the execution state to the target boundary and resumes execution as a new branch. The erroneous segment is therefore excluded from the active path, while the valid progress before the target boundary is preserved, isolating the error from subsequent decisions. As shown in Figure~\ref{fig:pr_tokens}, \ourmethod occupies the optimal upper-left quadrant, achieving stronger task progress with lower token consumption.


Our contributions are as follows. 
(1) We propose \ourmethod, which organizes agentic memory as a two-layer hierarchical tree whose root-to-current path provides a complete execution state by construction, shifting memory from similarity-driven retrieval to compact execution-state management.
(2) We design four coupled operations that make this tree an agent-manipulable object, forming a closed-loop state-management cycle that isolates errors into separate branches and keeps the active execution state free from erroneous traces.
(3) On MemoryArena, \ourmethod improves the task success rate by $7.8$--$20.4$ percentage points over baselines on average, while reducing token consumption by 55.1\% compared with the full-context approach.
\section{Background and Related Work}
\label{sec:related}

\subsection{Problem Setting}
\label{sec:problem}

Long-horizon agent tasks with interdependent decisions can be formulated as a Markov decision process (MDP)~\citep{markovian, yao2023react}. At step $t$, the environment state $s_t \in \mathcal{S}$ evolves deterministically as $s_{t+1}=T(s_t,a_t)$ after action $a_t \in \mathcal{A}$, and the agent receives an observation $o_t$ describing the resulting state. As a result, the interaction history is $h_t=(a_1,o_1,\ldots,a_t,o_t)$. As $t$ grows, this history can exceed the model's effective context window~\citep{memgpt2024,generativeagents,lostinthemiddle2024,reflexion2023}, making it the central challenge to organize $h_t$ compactly while still supporting complete state reconstruction.

This formulation highlights two requirements. First, since $s_t$ is determined by previous actions $(a_1,\ldots,a_t)$, a sufficient memory representation must preserve the decision chain on which each step depends rather than only relevant entries. Second, if an action $a_k$ is erroneous, downstream states $s_{k+1},\ldots,s_t$ may become invalid; recovery therefore requires identifying the error origin, reverting to $s_k$, and re-executing from that point. This distinguishes our setting from traditional memory benchmarks~\citep{locomo2024, longmemeval2024, amabench2026, meme2026}, which mainly test recall of facts, preferences, or events from past conversations or traces. Since answers in these benchmarks do not alter the environment or invalidate future states, they measure retrieval fidelity rather than dynamic execution-state management.

\subsection{Memory and Retrieval Systems for Agents}
\label{sec:related_memory}

A natural approach to managing long histories is retrieval-augmented generation (RAG), which augments the LLM context with information retrieved from an external store. Existing RAG methods include \textit{direct retrieval} with sparse or dense matching~\citep{bm25, rag2020, realm2020, dpr2020}, \textit{iterative retrieval} with query refinement~\citep{retro2022, queryrewrite2023}, \textit{graph-structured RAG} for multi-hop reasoning~\citep{graphrag2024, hipporag2024, hipporag2025}, and \textit{memory-augmented RAG} that uses a lightweight model to form global memory or retrieval clues~\citep{memorag2024}. These methods are effective for grounding generation in external knowledge, but the retrieved corpus is typically static and independent of the agent's action-conditioned state.

Agent memory systems instead store the agent's evolving history. They differ in storage design: \textit{flat} systems keep independent records retrieved by similarity~\citep{mem02025, simplemem2025, amem2025, nemori2025}; \textit{graph-based} systems organize memories through entity or event relations~\citep{zep2025, telemem2026, mragent2026, memorymattermore2026, AriadneMem2026}; \textit{hierarchical} systems maintain multiple granularities to balance detail and compression~\citep{memgpt2024, memoryos2025, evermemos2026, himem2026, timem2026}; and \textit{hybrid} systems combine granularities or narrative structures for compact coverage~\citep{memweaver2026, engram2025, omem2025, amory2026}. Other work improves retrieval with prospective indexing or retrospective reflection~\citep{rmm2025, hindsight2025, cma2026}.

Despite this diversity, these systems commonly expose memory through similarity-driven update and retrieval: they maintain a store $\mathcal{M}$ via $\mathcal{M}\leftarrow\texttt{Update}(\mathcal{M},(a_t,o_t))$ and retrieve entries $\texttt{Retrieve}(\mathcal{M},q)\to\{e_1,\ldots,e_k\}$ by semantic relevance to query $q$. This design reduces context length but does not preserve the path structure needed for long-horizon tasks with interdependent decisions. It therefore causes \textbf{state fragmentation}, where the decision chain defining $s_t$ is scattered across semantic fragments, and insufficient \textbf{error isolation}, where valid and erroneous trajectories coexist in the same memory pool without structural boundaries for rollback. Figure~\ref{fig:case_study_baseline_failure} illustrates both failures on shopping tasks in MemoryArena~\citep{memoryarena2026}.

\section{Method}
\label{sec:method}

\begin{figure*}[t]
\centering
\includegraphics[width=.9\textwidth]{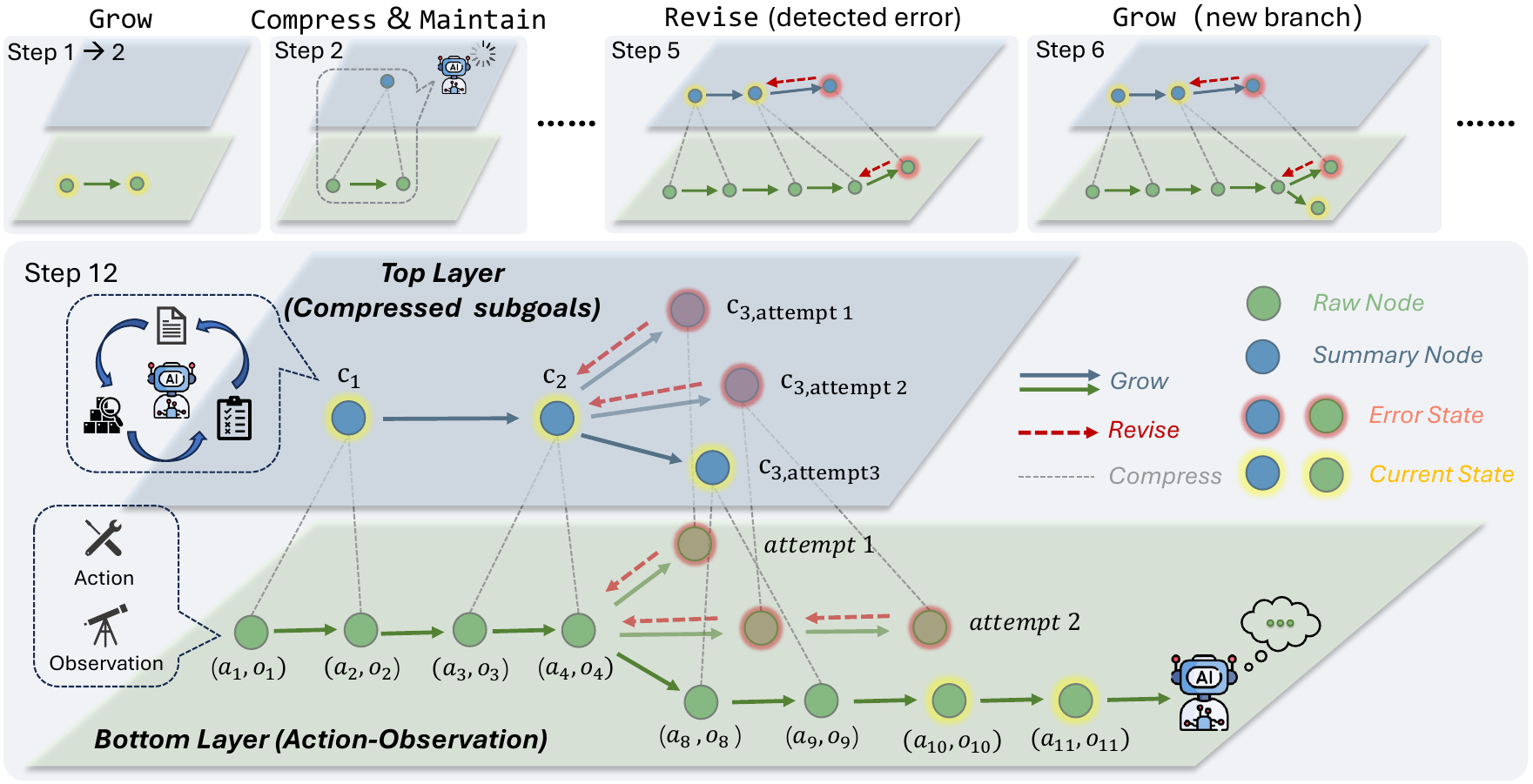}
\caption{\textbf{Overview of \ourmethod}. \ourmethod maintains a two-layer execution-state tree: raw action-observation nodes grow in the bottom layer, while completed subgoals are compressed to the top layer. When an error is detected, \texttt{Revise} restores the target boundary and resumes exploration along a new branch, preserving unaffected progress.}
\label{fig:mage_overview}
\end{figure*}

To address the issues inherent in similarity-driven memory systems, we propose \ourmethod, which shifts agentic memory from passive semantic storage and retrieval to active execution state management. Figure~\ref{fig:mage_overview} illustrates the overall design.

\subsection{Overview}
\label{sec:overview}

Cognitive science suggests that humans performing complex sequential tasks rely on coordinated neural mechanisms. The prefrontal cortex organizes behavior into hierarchical subgoals and chunks completed segments to free working memory for subsequent planning~\citep{botvinick2008}. The anterior cingulate cortex monitors execution and signals failures at subgoal boundaries before they propagate to downstream decisions~\citep{yeung2004}. After detecting errors, executive control selectively backtracks to the relevant boundary and repairs the affected segment while preserving unaffected goal structure~\citep{duncan2001}. This cycle of chunking, monitoring, and correction motivates a memory system that manages execution state actively rather than merely storing past information.

Motivated by this architecture, we propose \ourmethod (\textbf{M}emory as \textbf{A}gent-\textbf{G}uided \textbf{E}xploration), which represents the agent's execution history as a two-layer hierarchical state tree. The bottom layer records raw action-observation nodes in execution order, preserving fine-grained state dependencies. The top layer stores summary nodes that cover completed bottom-layer segments, progressively chunking long local traces into compact subgoal-level states. Together, these two layers keep the root-to-current path complete rather than fragmented, while bounding the context and retaining the boundaries needed for future revision. Based on this tree, \ourmethod constructs the agent-facing execution state $\mathcal{S}$, consisting of compressed summaries $\mathcal{C}$, recent raw trace $\mathcal{R}$, and execution hints $\mathcal{H}$ from previously explored branches and diagnostic notes.

Given this state representation, we design four operations to maintain the tree and refresh $\mathcal{S}$ in a closed loop, mirroring the cognitive cycle summarized in Table~\ref{tab:cognitive_mapping}. During execution, \texttt{Grow} appends each new action-observation pair to the bottom-layer tree, extending the recent raw trace $\mathcal{R}$. \texttt{Compress} moves completed raw segments from $\mathcal{R}$ into top-layer summaries in $\mathcal{C}$, freeing context while preserving subgoal boundaries. \texttt{Maintain} acts as a boundary-level error monitor, validating each new summary before it becomes trusted memory and recording diagnostic notes. Upon detecting an error, \texttt{Revise} provides selective correction by restoring $\mathcal{C}$ and $\mathcal{R}$ to the relevant boundary, injecting diagnostic feedback into $\mathcal{H}$, and resuming execution as a new branch from that point.

\begin{table}[t]
\centering
\small
\caption{\ourmethod operations parallel the cognitive mechanisms underlying human complex task execution.}
\label{tab:cognitive_mapping}
\resizebox{\columnwidth}{!}{%
\begin{tabular}{@{}lll@{}}
\toprule
\textbf{Cognitive Mechanism} & \textbf{Function} & \textbf{Operation} \\
\midrule
\makecell[l]{Hierarchical chunking\\\citep{botvinick2008}} & \makecell[l]{Organize subgoals;\\free working memory} & \makecell[l]{\texttt{Grow} +\\\texttt{Compress}} \\
\addlinespace
\makecell[l]{Error monitoring\\\citep{yeung2004}} & \makecell[l]{Detect errors at\\subgoal boundaries} & \texttt{Maintain} \\
\addlinespace
\makecell[l]{Selective correction\\\citep{duncan2001}} & \makecell[l]{Backtrack to error origin;\\correct affected branch} & \texttt{Revise} \\
\bottomrule
\end{tabular}%
}
\end{table}

This closed-loop design directly addresses state fragmentation and ineffective error isolation. First, \ourmethod constructs the current execution state from one path of the tree, combining compressed summaries for completed segments with raw traces for recent steps instead of retrieving disconnected memory entries based on similarity. Second, boundary-level maintenance and revision prevent erroneous segments from entering or contaminating the active execution state. When an error is detected, \texttt{Revise} branches from the target boundary, isolating the affected segment while preserving valid progress elsewhere. We next present the hierarchical execution state tree and the four state-transition operations.

\subsection{Hierarchical Execution State Tree}
\label{sec:tree}

\ourmethod organizes the agent's execution history as a two-layer hierarchical tree with a unified node structure (Table~\ref{tab:node_struct}). The bottom layer records every raw action-observation pair as a node, preserving the fine-grained state of the execution. The root-to-current path through this layer yields the complete execution trajectory, while children of each node expose previously explored alternatives that can help the agent avoid repeating errors. When the agent revises a decision, new actions branch as siblings of the failed path, structurally isolating erroneous traces from valid ones. The top layer compresses contiguous bottom-layer segments into summary nodes, bounding the context as the task progresses. Each top-layer node corresponds to a completed subgoal, and we apply \texttt{Maintain} and \texttt{Revise} at these subgoal boundaries, the same locus where the prefrontal cortex chunks completed segments and the anterior cingulate cortex monitors for errors before they propagate (Table~\ref{tab:cognitive_mapping}).

\begin{table}[t]
\centering
\small
\caption{Node structure of the execution state tree.}
\label{tab:node_struct}
\begin{tabular}{@{}lp{0.68\columnwidth}@{}}
\toprule
\textbf{Field} & \textbf{Description} \\
\midrule
\texttt{id} & Unique identifier \\
\texttt{content} & Action-observation pair (bottom layer) or compressed summary (top layer) \\
\texttt{parent} & Pointer to parent node \\
\texttt{children} & Set of child node pointers \\
\texttt{cover\_nodes} & Ordered bottom-node pointers covered by this summary (top layer only) \\
\texttt{note} & Diagnostic feedback (top layer only) \\
\bottomrule
\end{tabular}
\end{table}

At runtime, \ourmethod navigates this two-layer structure with pointers $p_b$ and $p_t$, tracking the agent's current positions in the bottom and top layers, respectively, and a global step $\ell$ assigning monotonically increasing ids to newly created nodes.

Based on the hierarchical tree, \ourmethod derives the agent's execution state $\mathcal{S}$ by composing three parts: (1) \textit{compressed state} $\mathcal{C}$, consisting of top-layer summaries along the root-to-$p_t$ path, each annotated by its step id, allowing the agent to revise failed subgoals from the corresponding boundary upon detecting errors; (2) \textit{raw state} $\mathcal{R}$, the bottom-layer nodes accumulated since the last compression that provide fine-grained recent context; and (3) \textit{execution hint} $\mathcal{H}$, which surfaces children of $p_b$ and $p_t$ to reveal previously explored alternatives, along with diagnostic feedback from prior failed attempts. This representation equips the agent with a complete execution state and corrective guidance, sustaining coherent decision-making over long-horizon tasks with interdependent steps.

\subsection{State-Transition Operations}
\label{sec:operations}

The hierarchical tree becomes an active execution manager through four operations that transition the execution state as the agent progresses. Algorithm~\ref{alg:operations} in Appendix provides the pseudocode of these operations.

\paragraph{Grow.}
When the agent executes an action and receives an observation, \ourmethod automatically invokes \texttt{Grow} to update the bottom layer. If $p_b$ already has a child with identical content from a prior exploration, the pointer advances to that child, merging back into the explored path without duplication; otherwise, a new node is created and linked as a child of $p_b$:
\[
\resizebox{\columnwidth}{!}{$
\begin{aligned}
p_b' =
\begin{cases}
c, \quad \exists c\in p_b.\text{children}: c.\text{content}=(a,o),\\
\operatorname{NewNode}(\ell,(a,o)),\ \ell\leftarrow\ell+1, \quad \text{otherwise}.
\end{cases}
\end{aligned}
$}
\]
The raw state $\mathcal{R}$ is then extended with the new action-observation pair, and the execution hint $\mathcal{H}$ is updated with children of the current node:
\[
\mathcal{R}' = \mathcal{R}\Vert(a,o),\quad
\mathcal{H}' = \operatorname{Update}(\mathcal{H},p_b'.\text{children})
\]
This informs the agent of continuations attempted in prior explorations and helps it avoid repeating failed strategies.

\paragraph{Compress.}
\texttt{Compress} bounds context growth by replacing a completed bottom-layer segment with a top-layer summary node, freeing space while preserving the decision boundary needed for later recovery. It is invoked when the agent marks a subgoal complete with summary content provided as an argument, or by \ourmethod as a fallback when the raw state $\mathcal{R}$ exceeds a length threshold. Through this boundary-aware compression, \ourmethod avoids interrupting unfinished subgoals and keeps the state compact without discarding dependencies needed by future decisions.

Operationally, \texttt{Compress} traces the bottom-layer tree from $p_b$ back to the last compressed boundary (recorded as $p_t.\text{cover\_nodes}[-1]$), and uses the traversed nodes in execution order as the new summary node's $\text{cover\_nodes}$. If a child of $p_t$ already covers the same bottom nodes, it is reused; otherwise, a new summary node is created and inserted into the top-layer tree:
\[
\resizebox{\columnwidth}{!}{$
\begin{aligned}
C_b &= \operatorname{Trace}(p_t.\text{cover\_nodes}[-1],p_b),\\
\ell_b &= p_t.\text{cover\_nodes}[-1].\text{id},\\
p_t' &=
\begin{cases}
c, \quad \exists c\in p_t.\text{children}: c.\text{cover\_nodes}=C_b,\\
\operatorname{NewNode}(\ell_b,\text{sum\_content},C_b), \quad \text{otherwise}.
\end{cases}
\end{aligned}
$}
\]
Then, \texttt{Compress} clears the current raw state $\mathcal{R}$, appends the summary content to the compressed state $\mathcal{C}$, and updates the execution hint $\mathcal{H}$ with children of $p_t'$:
\[
\resizebox{\columnwidth}{!}{$
\mathcal{R}'=\emptyset,\quad
\mathcal{C}'=\mathcal{C}\Vert(p_t'.\text{content}),\quad
\mathcal{H}'=\operatorname{Update}(\mathcal{H},p_t'.\text{children}).
$}
\]
This exposes previously attempted subgoals from the new boundary while keeping the compressed state compact.

\paragraph{Maintain.}
Immediately after compression, \texttt{Maintain} validates the just-completed subgoal before the new summary becomes a trusted part of memory. This check protects the execution state from incorrect memory writes, allowing \ourmethod to detect missing information, unsatisfied task requirements, or broken dependencies before such errors accumulate. An LLM examines the compressed subtree together with the summary content and task instruction:
\[
f=\operatorname{LLM}(\text{task\_inst},\ p_t.\text{cover\_nodes},\ p_t.\text{content}).
\]
If validation passes, execution continues. Otherwise, \texttt{Maintain} records the diagnostic feedback $f$ in $p_t.\text{note}$ and returns a failure signal with the revision target $p_t.\text{id}$.

\paragraph{Revise.}
Triggered by a \texttt{Maintain} failure or invoked proactively by the agent upon detecting an error, \texttt{Revise} restores the active path to the target step $\ell_t$, which is either returned by \texttt{Maintain} or selected from exposed compressed-state boundaries. \ourmethod rolls both pointers backward until the target is reached, where re-exploration begins:
$$
(p_t',p_b')=\operatorname{Restore}(p_t, p_b, \ell_t).
$$
The compressed state $\mathcal{C}$ and raw state $\mathcal{R}$ are reverted to these positions, while the execution hint $\mathcal{H}$ is updated with diagnostic feedback and alternatives from the restored nodes, providing extra guidance that helps avoid repeating the same error:
\[
\resizebox{\columnwidth}{!}{$
\begin{aligned}
\mathcal{C}'&=\operatorname{RestoreState}(p_t'),\quad \mathcal{R}'=\emptyset,\\
\mathcal{H}'&=\operatorname{Update}(\mathcal{H},p_t'.\text{children}\cup p_b'.\text{children}\cup\{f\}).
\end{aligned}
$}
\]
Subsequent actions branch from this restored point as sibling paths, achieving error isolation without discarding valid progress on other branches.

\section{Experiments}
\label{sec:experiments}

\subsection{Experimental Setup}
\label{sec:exp_setup}

\paragraph{Benchmark.} We evaluate on MemoryArena~\citep{memoryarena2026}, an interdependent long-horizon benchmark where agents operate in a continuous Memory-Agent-Environment loop for up to hundreds of steps. Unlike conventional benchmarks that test static fact retrieval or question answering over past dialogues and traces~\citep{locomo2024, longmemeval2024, amabench2026}, MemoryArena follows action-conditioned MDPs: each action can reshape future constraints, so success requires tracking the evolving execution state rather than recalling facts. 

MemoryArena spans four domains with long dependency structures: \textit{Bundled Web Shopping}~\citep{webshop2022}, where the agent purchases a bundle of related products and later choices depend on earlier items; \textit{Group Travel Planning}~\citep{travelplanner2024}, in which the agent coordinates multi-person itineraries to satisfy interdependent preferences; \textit{Progressive Web Search}~\citep{chen2025browsecomp}, where the model answers complex queries progressively using information gathered from previous sub-queries; and \textit{Formal Reasoning}, in which the agent proves complex claims through sequential derivations that build on previously established results.

To further evaluate the generalizability of \ourmethod, we provide additional results on another long-horizon benchmark, DeepPlanning~\citep{deepplanning2026}, in Appendix~\ref{app:deepplanning}. Since the results exhibit trends consistent with those on MemoryArena, we focus on MemoryArena in the main text for clarity.

\paragraph{Baselines.} We compare \ourmethod against representative methods across three paradigms. \textit{Full Context} retains full interaction history. \textit{RAG systems} include HippoRAG2~\citep{hipporag2025}, which builds a knowledge graph and applies Personalized PageRank for multi-hop retrieval, and MemoRAG~\citep{memorag2024}, which uses a lightweight memory model to generate retrieval clues. \textit{Memory systems} include Mem0~\citep{mem02025}, which extracts and consolidates facts into graph-based memory, ReasoningBank~\citep{reasoningbank2025}, which distills reusable reasoning strategies from past experiences, MemoryOS~\citep{memoryos2025}, which maintains hierarchical storage layers with dynamic cross-level updating, and SimpleMem~\citep{simplemem2025}, which performs semantic compression and recursive consolidation for efficient memory management. All methods use the default hyperparameters from their original papers.

\paragraph{Model.} All methods use Qwen3.6-27B~\citep{qwen3.6-27B} as the backend LLM with a ReAct~\citep{yao2023react} loop for agent exploration. Baselines requiring dense embeddings use Qwen3-8B-Embedding~\citep{qwen3-embedding}. Inference is performed on NVIDIA A100 GPUs~\citep{A100} using vLLM~\citep{kwon2023efficient} 0.20.0 with Python 3.12. Results with two additional backend models, Qwen3.6-35B-A3B~\citep{qwen3.6-35B-A3B} and Gemma4-31B-it~\citep{gemma4-31B}, are reported in Appendix~\ref{app:backend_models}. As the results exhibit consistent trends across models, we use Qwen3.6-27B as the representative backend model for the main-text analysis.

\paragraph{Metrics.} Following MemoryArena~\citep{memoryarena2026}, we report average Task Success Rate (SR), Task Progress Score (PS), and total token consumption. SR measures full task completion: all subtasks must be correct in Shopping and Travel Planning, while the final subtask determines success in the other two domains. PS measures completed-subtask fraction, and token consumption includes both prompt and generation tokens.

\subsection{Main Results}
\label{sec:main_results}

\begin{table*}[t]
\centering
\caption{\textbf{Main results on MemoryArena (Qwen3.6-27B).} SR = Task Success Rate; PS = Task Progress Score; \#tokens = average token consumption per task. Best results in \textbf{bold}, second best \underline{underlined}. \ourmethod achieves the best task accuracy while reducing token consumption.}
\label{tab:main}
\resizebox{\textwidth}{!}{
\setlength{\tabcolsep}{5pt}
\begin{tabular}{l ccc @{\hspace{14pt}} ccc @{\hspace{14pt}} ccc @{\hspace{14pt}} ccccc}
\toprule
& \multicolumn{3}{c@{\hspace{14pt}}}{\multirow{2}[2]{*}{\makecell{\textbf{Bundled}\\\textbf{Web Shopping}}}} & \multicolumn{3}{c@{\hspace{14pt}}}{\multirow{2}[2]{*}{\makecell{\textbf{Group}\\\textbf{Travel Planning}}}} & \multicolumn{3}{c@{\hspace{14pt}}}{\multirow{2}[2]{*}{\makecell{\textbf{Progressive}\\\textbf{Web Search}}}} & \multicolumn{5}{c}{\textbf{Formal Reasoning}} \\
\cmidrule(lr){11-15}
& & & & & & & & & & \multicolumn{2}{c}{Math} & \multicolumn{2}{c}{Physics} & ~ \\
\midrule
\textbf{Method} & SR & PS & \cellcolor{blue!5}\#tokens & SR & PS & \cellcolor{blue!5}\#tokens & SR & PS & \cellcolor{blue!5}\#tokens & SR & PS & SR & PS & \cellcolor{blue!5}\#tokens \\
\midrule
Full Context & \underline{0.3333} & \underline{0.7578} & \cellcolor{blue!5}\textit{1528K} & 0.0519 & 0.4070 & \cellcolor{blue!5}\textit{3211K} & \underline{0.4842} & \underline{0.2803} & \cellcolor{blue!5}\textit{6045K} & 0.4000 & 0.4124 & \textbf{0.6500} & 0.6977 & \cellcolor{blue!5}\textit{1782K} \\
\midrule
HippoRAG2 & 0.2067 & 0.7189 & \cellcolor{blue!5}\textit{1720K} & \underline{0.0963} & \textbf{0.5569} & \cellcolor{blue!5}\textit{3153K} & 0.3620 & 0.2133 & \cellcolor{blue!5}\textit{2865K} & \textbf{0.4500} & 0.4153 & \underline{0.6000} & \underline{0.7093} & \cellcolor{blue!5}\textit{2268K} \\
MemoRAG & 0.2067 & 0.7200 & \cellcolor{blue!5}\textit{1251K} & 0.0481 & 0.4731 & \cellcolor{blue!5}\textit{2829K} & 0.4208 & 0.2535 & \cellcolor{blue!5}\textit{3535K} & \textbf{0.4500} & 0.4237 & \textbf{0.6500} & \textbf{0.7209} & \cellcolor{blue!5}\textit{1314K} \\
\midrule
Mem0 & 0.1933 & 0.6822 & \cellcolor{blue!5}\textit{1753K} & 0.0259 & 0.3498 & \cellcolor{blue!5}\textit{2786K} & 0.3529 & 0.2206 & \cellcolor{blue!5}\textit{2715K} & 0.4000 & 0.4040 & 0.5500 & 0.6977 & \cellcolor{blue!5}\textit{1853K} \\
ReasoningBank & 0.1133 & 0.6033 & \cellcolor{blue!5}\textit{868K} & 0.0000 & 0.2346 & \cellcolor{blue!5}\textit{1463K} & 0.3032 & 0.1993 & \cellcolor{blue!5}\textit{1923K} & \underline{0.4250} & 0.4266 & 0.5500 & 0.6744 & \cellcolor{blue!5}\textit{1187K} \\
MemoryOS & 0.2000 & 0.7044 & \cellcolor{blue!5}\textit{1448K} & 0.0259 & 0.4405 & \cellcolor{blue!5}\textit{2984K} & 0.3303 & 0.2084 & \cellcolor{blue!5}\textit{2973K} & \textbf{0.4500} & 0.4209 & 0.5500 & 0.6977 & \cellcolor{blue!5}\textit{1452K} \\
SimpleMem & 0.1600 & 0.6767 & \cellcolor{blue!5}\textit{2939K} & 0.0222 & 0.3246 & \cellcolor{blue!5}\textit{4616K} & 0.3575 & 0.2310 & \cellcolor{blue!5}\textit{3715K} & 0.4000 & \underline{0.4294} & 0.5500 & 0.6628 & \cellcolor{blue!5}\textit{1656K} \\
\midrule
\textbf{\ourmethod (Ours)} & \textbf{0.3933} & \textbf{0.7778} & \cellcolor{blue!5}\textit{1015K} & \textbf{0.1519} & \underline{0.5351} & \cellcolor{blue!5}\textit{1978K} & \textbf{0.5656} & \textbf{0.3790} & \cellcolor{blue!5}\textit{1727K} & \underline{0.4250} & \textbf{0.4492} & \textbf{0.6500} & 0.6977 & \cellcolor{blue!5}\textit{1195K} \\
\bottomrule
\end{tabular}
}
\end{table*}

Table~\ref{tab:main} presents task performance across four domains. On tasks with complex state dependencies, RAG and memory-based baselines underperform the full-context approach, with SR drops of $12.7$--$22.0$~pp on Web Shopping and $6.3$--$18.1$~pp on Web Search. This gap suggests that similarity-driven retrieval fragments the execution state into isolated entries, discarding structural dependencies retained in full interaction history. The partial exceptions occur when dependencies can be recovered as sparse local evidence. In Formal Reasoning, baselines perform comparably or slightly better than the full-context approach because mathematical dependencies are explicit and sparse, enabling precise lemma retrieval. Similarly, HippoRAG2 achieves good results on Travel Planning because local constraints are anchored by stable entities and attributes (e.g., hotel--city, restaurant--cuisine), making graph retrieval effective for recovering candidate facts; yet full travel plans require cross-traveler constraints from prior decisions, and its fragmented retrieved triples do not preserve this evolving execution state, so its PS advantage does not translate into higher SR than \ourmethod.

Conversely, \ourmethod outperforms the full-context approach by average margins of $7.8$~pp in SR and $8.7$~pp in PS. This gain stems from treating memory as an active execution-state manager, not a passive archive. Modeling the trajectory as an active path within a two-layer hierarchical tree, \ourmethod maintains a coherent current state, reuses historical exploration to avoid recurring errors, and quarantines flawed segments into inactive branches to prevent contamination of the active path.

\ourmethod achieves substantial gains when execution state management is the key bottleneck. For example, Shop, Travel, and Search tasks involve long trajectories (82--237 steps; see Table~\ref{tab:dataset_statistics}) with evolving dependencies, where effective memory management is particularly important. In domains where model reasoning capacity, rather than execution state management, is the primary bottleneck, such as Math and Physics with shorter traces (13.8--24.7 steps), \ourmethod still achieves comparable performance. This suggests that \ourmethod does not introduce adverse effects when memory is less critical.

\subsection{Token Efficiency}
\label{sec:token_efficiency}

Regarding efficiency, Table~\ref{tab:main} also shows token consumption per task. While RAG and memory baselines theoretically reduce context length, this benefit primarily materializes in document-heavy environments like Web Search, where they reduce token usage by 38.5--68.2\% compared with the full-context approach. In domains with shorter observations and frequent state updates, the overhead of memory maintenance (e.g., extraction, query rewriting), compounded by the additional reasoning and execution steps required to synthesize fragmented retrieved states, frequently eclipses the compression savings. Consequently, systems like HippoRAG2 and Mem0 end up consuming 12.6--14.7\% more tokens than the full-context approach on Web Shopping. In contrast, \ourmethod consistently reduces token usage by 32.9--71.4\% across all domains. Unlike traditional memory systems that require continuous, token-heavy auxiliary LLM calls to extract entities or generate queries, \ourmethod maintains the tree structure deterministically and invokes auxiliary LLMs only during \texttt{Compress} and \texttt{Maintain} at natural subgoal boundaries. This design minimizes maintenance overhead without compromising state integrity.

\subsection{Ablation Study}
\label{sec:ablation}

\begin{table}[t]
\centering
\caption{\textbf{Ablation study results on MemoryArena (Qwen3.6-27B).} Each row removes one mechanism from the full \ourmethod.}
\label{tab:ablation}
\resizebox{\columnwidth}{!}{
\begin{tabular}{l ccc ccc}
\toprule
\multicolumn{1}{c}{\multirow{2}[2]{*}{\textbf{Variant}}} & \multicolumn{3}{c}{\textbf{Web Shopping}} & \multicolumn{3}{c}{\textbf{Travel Planning}} \\
\cmidrule(lr){2-4} \cmidrule(lr){5-7}
& SR & PS & \cellcolor{blue!5}\#tokens & SR & PS & \cellcolor{blue!5}\#tokens \\
\midrule
\ourmethod (Full) & 0.3933 & 0.7778 & \cellcolor{blue!5}\textit{1015K} & 0.1519 & 0.5351 & \cellcolor{blue!5}\textit{1978K} \\
\quad w/o Compress & 0.3200 & 0.7233 & \cellcolor{blue!5}\textit{2469K} & 0.0852 & 0.4496 & \cellcolor{blue!5}\textit{3539K} \\
\quad w/o Maintain & 0.3267 & 0.7389 & \cellcolor{blue!5}\textit{887K} & 0.1000 & 0.4683 & \cellcolor{blue!5}\textit{1256K} \\
\quad w/o Revise & 0.3533 & 0.7378 & \cellcolor{blue!5}\textit{1157K}  & 0.1000 & 0.4708 & \cellcolor{blue!5}\textit{2551K} \\
\bottomrule
\end{tabular}
}
\end{table}

To validate the contribution of the main state-management mechanisms, we evaluate \ourmethod variants that remove one mechanism at a time. Table~\ref{tab:ablation} shows that all three mechanisms contribute to reliable execution-state management. 

Removing \texttt{Compress} consistently hurts task completion, reducing SR by $7.3$~pp on Web Shopping and $6.7$~pp on Travel Planning, while increasing token consumption to $2.4\times$ and $1.8\times$ that of the full model. This confirms that simply retaining the raw action-observation stream is not a sufficient substitute for memory: without boundary-aware compression, the active state becomes diluted by low-level traces, and \texttt{Maintain} must verify an increasingly long trajectory.

Removing \texttt{Maintain} leads to a different failure mode. Although it reduces token usage by avoiding boundary-level verification, SR drops by $5.2$--$6.7$~pp on two domains. This result demonstrates that memory writes should be validated before they become trusted state. Otherwise, incomplete or erroneous subgoal summaries are committed to the active path and later decisions are conditioned on corrupted execution state.

Finally, removing \texttt{Revise} lowers SR by $4.0$--$5.2$~pp on two domains. This shows that error detection alone is insufficient: after a failed boundary is identified, the agent must restore the corresponding state, branch away from the flawed segment, and continue from the preserved valid prefix; otherwise, the error remains on the active path and contaminates later decisions.

Notably, these weakened variants remain competitive with or superior to the baselines in Table~\ref{tab:main}. This suggests that organizing memory around the execution path mitigates the state fragmentation caused by semantic retrieval, while the full operation loop further prevents corrupted or failed segments from contaminating the active state.

\subsection{Trace Breakdown Analysis}

To further understand the effect of \ourmethod's four state-transition operations, we break down the traces of the \textit{Group Travel Planning} tasks and analyze the frequency of each operation. 

We find that only $\sim$3.7\% of the total steps invoke \texttt{Compress} and \texttt{Maintain}, while $\sim$1.2\% invoke \texttt{Revise}. An additional $\sim$6.0\% of steps correspond to re-execution after \texttt{Revise}, leaving the majority ($\sim$89.1\%) to normal task execution. This breakdown shows that the auxiliary operations introduce only limited overhead. Moreover, after \texttt{Revise}, approximately 67.2\% of errors are successfully corrected, demonstrating its effectiveness in mitigating errors from propagating to downstream branches. The remaining errors are primarily due to the model's reasoning limitations and task complexity, which are beyond the scope of memory management. 
Overall, these results show that \ourmethod introduces minimal overhead while substantially improving execution-state management in interdependent long-horizon tasks.

\section{Conclusion}
\label{sec:conclusion}
We presented \ourmethod, a memory framework that reframes long-horizon agent memory as active execution-state management rather than similarity-driven retrieval. By organizing interaction history as a two-layer hierarchical state tree, \ourmethod preserves the active root-to-current execution path while compressing completed subgoals and exposing execution hints from previously explored branches. Its four coupled operations allow agents to extend traces, bound context growth, validate newly compressed states, and isolate erroneous segments through branching. Experiments on MemoryArena show that this design improves task success across diverse long-horizon domains while substantially reducing token consumption. These results suggest that preserving execution-state structure is a key principle for building reliable and efficient memory systems for real-world LLM agents.

\section*{Limitations}
\label{sec:limitations}
\ourmethod operates through prompting without task-specific training. Learning dedicated models for state compression and error detection could further improve the quality of execution-state management, which we leave for future work. Additionally, while we demonstrate the effectiveness of our approach on long-horizon tasks with interdependent decisions, extending \ourmethod to broader scenarios such as multimodal interaction or collaborative multi-agent workflows remains an important direction for future work.

\section*{Ethical Considerations}
This work studies memory mechanisms for LLM-based agents and is not tied to any specific high-stakes deployment. Nevertheless, improving long-horizon agent reliability may also benefit dual-use applications, such as automated information gathering or tool-use workflows that could be misused if deployed without safeguards. In practice, execution histories may contain sensitive information, so systems based on \ourmethod should apply data minimization, access control, retention limits, and domain-specific safety checks when storing or reusing traces. Since \ourmethod does not train new models, its direct environmental impact primarily comes from inference-time computation.

\section*{Acknowledgments}
We thank all anonymous reviewers and area chairs for their insightful comments.
This work is supported in part by the National Major Scientific and Technological Project for New Generation Artificial Intelligence under Grant No.2025ZD0122104.

\bibliography{ref}

\newpage
\appendix
\section{Pseudocode of Four State-Transition Operations}
\label{sec:algorithms}

Algorithm~\ref{alg:operations} provides the full pseudocode for the four state-transition operations described in Section~\ref{sec:operations}, i.e., \texttt{Grow}, \texttt{Compress}, \texttt{Maintain}, and \texttt{Revise}.

\begin{algorithm}[H]
\caption{State-Transition Operations of \ourmethod}
\label{alg:operations}
\footnotesize
\algrenewcommand\alglinenumber[1]{\scriptsize\textcolor{gray}{\textup{#1:}}}
\algrenewcommand\algorithmiccomment[1]{\hfill\textcolor{gray}{$\triangleright$ #1}}
\algrenewcommand\algorithmicrequire{\textbf{Global Variable:}}
\algrenewcommand\algorithmicensure{\textbf{Execution State:}}
\begin{algorithmic}[1]
\Require Pointers to the bottom-layer tree $p_b$ and top-layer tree $p_t$, global step $\ell$.
\Ensure $\mathcal{S}=(\mathcal{C},\mathcal{R},\mathcal{H})$ for compressed state, raw state, and execution hint.

\Statex
\Function{Grow}{$a,o$}
    \ForAll{$c\in p_b.\text{children}$}
        \If{$c.\text{content}=(a,o)$}  \Comment{merge node}
            \State $p_b \gets c$
            \State $\mathcal{R}\gets\mathcal{R}\mathbin{\Vert}(a,o)$
            \State $\mathcal{H}\gets\Call{Update}{\mathcal{H},p_b.\text{children}}$
            \State \Return
        \EndIf
    \EndFor
    \State $v \gets \Call{NewNode}{\ell,(a,o)}$; \quad $\ell \gets \ell+1$
    \State $v.\text{parent} \gets p_b$; \quad $p_b.\text{children}\gets p_b.\text{children}\cup\{v\}$
    \State $p_b \gets v$; \quad $\mathcal{R}\gets\mathcal{R}\mathbin{\Vert}(a,o)$
    \State $\mathcal{H}\gets\Call{Update}{\mathcal{H},p_b.\text{children}}$
\EndFunction

\Statex
\Function{Compress}{$m$} \Comment{input summary content}
    \State $b \gets p_t.\text{cover\_nodes}[-1]$ \Comment{Compressed boundary}
    \State $C_b \gets \Call{Trace}{b, p_b}$ \Comment{track nodes in execution order}
    \ForAll{$c\in p_t.\text{children}$}  \Comment{merge node}
        \If{$c.\text{cover\_nodes}=C_b$}
            \State $c.\text{content} \gets m$; \quad $p_t \gets c$
            \State $\mathcal{R}\gets\emptyset$; \quad $\mathcal{C}\gets\mathcal{C}\mathbin{\Vert}(p_t.\text{content},p_t.\text{id})$
            \State $\mathcal{H}\gets\Call{Update}{\mathcal{H},p_t.\text{children}}$
            \State \Return
        \EndIf
    \EndFor
    \State $v \gets \Call{NewNode}{b.\text{id},m,C_b}$
    \State $v.\text{parent}\gets p_t$; \quad $p_t.\text{children}\gets p_t.\text{children}\cup\{v\}$
    \State $p_t\gets v$
    \State $\mathcal{R}\gets\emptyset$; \quad $\mathcal{C}\gets\mathcal{C}\mathbin{\Vert}(p_t.\text{content},p_t.\text{id})$
    \State $\mathcal{H}\gets\Call{Update}{\mathcal{H},p_t.\text{children}}$
\EndFunction

\Statex
\Function{Maintain}{$\tau$}  \Comment{input task instruction}
    \State $T \gets \Call{Flatten}{p_t.\text{cover\_nodes}}$ \Comment{flatten traces}
    \State $(q,f) \gets \Call{LLM}{\tau, T, p_t.\text{content}}$ \Comment{LLM judge}
    \If{$q$}
        \State \Return \textsc{Pass}
    \Else
        \State $p_t.\text{note}\gets f$
        \State \Return $(\textsc{Fail}, f, p_t.\text{id})$
    \EndIf
\EndFunction

\Statex
\Function{Revise}{$f,\ell_t$} \Comment{input feedback and target step}
    \While{$p_t.\text{id}\neq \ell_t$}
        \State $\mathcal{C}.\Call{Delete}{p_t}$; \quad $p_t\gets p_t.\text{parent}$
    \EndWhile
    \State $\mathcal{C}.\Call{Delete}{p_t}$
    \State $p_t\gets p_t.\text{parent}$ \Comment{skip the failed compressed node}
    \State $p_b\gets p_t.\text{cover\_nodes}[-1]$; \quad $\mathcal{R} \gets \emptyset$
    \State $\mathcal{H}\gets\Call{Update}{\mathcal{H},p_t.\text{children}\cup p_b.\text{children}\cup\{f\}}$
\EndFunction
\end{algorithmic}
\end{algorithm}

\section{Experiment Details}

\subsection{Prompt Templates}
\label{sec:prompts}

We provide the prompt templates used in our experiments across the four MemoryArena domains~\citep{memoryarena2026}. Each template specifies the task instruction and available action space that guide the agent's interaction with the environment.

\definecolor{bluex}{RGB}{41,98,168}
\definecolor{greenx}{RGB}{34,139,87}
\definecolor{orangex}{RGB}{196,120,32}

\begin{tcolorbox}[enhanced, breakable, boxrule=0.8pt, arc=4pt, fonttitle=\bfseries\small, fontupper=\small, before upper={\setlength{\parskip}{4pt}}, coltitle=white, colbacktitle=bluex, colback=bluex!3, colframe=bluex!70, left=4pt, right=4pt, top=4pt, bottom=4pt, title={Prompt for Bundled Web Shopping}]
{
You can interact with the webshop. Your goal is to purchase a bundle of items that are technically compatible and fit the budget. Each time you will receive one instruction with multiple options and you need to buy the best product among the options. For each option, search by the option text (do not miss any keywords) and browse multiple pages to find the most matched product, e.g., containing as many keywords mentioned in that option as possible, with the same weight, size, quantity, pack, brand, etc. Each option corresponds to only one best-matching product. The valid product may appear on later pages. Do NOT miss checking any option.

If given compatibility/avoid notes in the task, filter options/products by the compatibility notes (i.e., keywords of prior purchased products pair with keywords of the chosen product) and avoid notes (i.e., keywords of prior purchased products not pair with keywords of the chosen product). When applying compatibility/avoid notes, only consider exact keywords that appear in the text literally, do NOT infer the attributes of the product based on your internal knowledge. Compatibility and avoid notes should not be applied to the keywords of the chosen product itself, only apply them between prior purchased products and currently chosen product. If a product does not contain any keyword from the compatibility/avoid notes, it is compatible by default (no rule applies). Once you finish checking an option and find the corresponding product, store the information (product title, ASIN, price, rating if needed, etc) even the option is NOT compatible. Finally, compare all compatible products and navigate to the page of the product that satisfies the preference rule and buy it by clicking [Buy Now] button. You must buy exactly one product per task.

Constraint:
\vspace{-\parskip}
\begin{enumerate}[label=\arabic*., leftmargin=*, nosep, parsep=0pt]
	\item \textbf{Evaluate All:} Never only pick the first option you see; compare all candidates.
	\item \textbf{Total Budget:} All items combined must not exceed \$XX.
	\item \textbf{Product Search:} Search the product with the detailed description one by one. For example, use "search[Product A]" but not "search[Product A, Product B, Product C]".
	\item \textbf{Product Purchase:} You need to buy products on the order of the steps (i.e., Product 1 first, then Product 2, and so on).
\end{enumerate}
}
\end{tcolorbox}

\begin{tcolorbox}[enhanced, breakable, boxrule=0.8pt, arc=4pt, fonttitle=\bfseries\small, fontupper=\small, before upper={\setlength{\parskip}{4pt}}, coltitle=white, colbacktitle=bluex, colback=bluex!3, colframe=bluex!70, left=4pt, right=4pt, top=4pt, bottom=4pt, title={Prompt for Group Travel Planning}]
{
You are a travel planning agent that builds a multi-day itinerary for one traveler in a group by calling structured tools. Each task processes one traveler's query under shared group constraints; previously planned travelers are provided as context. You must search before selecting and only use option ids returned by search tools; never invent flights, restaurants, hotels, or attractions, and never construct option ids from names recalled in memory. Use the provided session id for every stateful action, and check cost and itinerary snapshots when budget or completeness is uncertain.

Planning strategy:
\vspace{-\parskip}
\begin{itemize}[leftmargin=*, nosep, parsep=0pt]
    \item For each day and slot (breakfast, lunch, dinner, attraction, accommodation), first search candidates, then select or book exactly one option from the returned ids.
    \item For relative constraints referencing another traveler, first recall the referenced slot from injected memory, convert to absolute search parameters, then search, verify each candidate against all conditions, and select the cheapest qualifying option.
\end{itemize}

Constraint:
\vspace{-\parskip}
\begin{enumerate}[label=\arabic*., leftmargin=*, nosep, parsep=0pt]
    \item \textbf{Grounded Selection:} Only select or book from ids returned by the most recent search; do not fabricate or reuse ids inferred from names in memory.
    \item \textbf{Group Dependencies:} Use injected memory and previously planned travelers to derive absolute search constraints for the current traveler.
    \item \textbf{Completion Criterion:} Call \texttt{Finish[done]} only when all required transportation legs are booked and every destination day has breakfast, lunch, dinner, attraction, and accommodation assigned.
\end{enumerate}
}
\end{tcolorbox}

\begin{tcolorbox}[enhanced, breakable, boxrule=0.8pt, arc=4pt, fonttitle=\bfseries\small, fontupper=\small, before upper={\setlength{\parskip}{4pt}}, coltitle=white, colbacktitle=bluex, colback=bluex!3, colframe=bluex!70, left=4pt, right=4pt, top=4pt, bottom=4pt, title={Prompt for Progressive Web Search}]
{
You are a deep research agent solving a complex question decomposed into progressive sub-questions. All sub-questions constrain the same entity. Answer each question by interacting with a search engine, i.e., searching the corpus and reading documents; never Finish from internal knowledge or lock onto a candidate without corpus evidence. Keep your reasoning process compact and focused.

Search strategy:
\vspace{-\parskip}
\begin{itemize}[leftmargin=*, nosep, parsep=0pt]
    \item Use focused 3-6 word queries. Try different search angles with synonyms, related concepts, broader/narrower terms, or different clue combinations.
    \item Do not repeat search by only reordering keywords. If a search keyword pollutes results, drop it. If no relevant results are found, rethink entity type, wording, region/culture, or assumptions.
    \item Each search may return multiple relevant documents; read several promising ones, not just the first.
    \item Store brief notes for failed searches and each document you read, whether relevant or not.
\end{itemize}

Constraint:
\vspace{-\parskip}
\begin{enumerate}[label=\arabic*., leftmargin=*, nosep, parsep=0pt]
    \item You must search the corpus and read a document before answering. Never use internal knowledge as your answer. Find it in documents first.
    \item Keep your reasoning process brief and output exactly one next action only; do not re-solve prior sub-questions, enumerate many candidates/search queries, or simulate observations.
    \item Return the answer exactly as it appears in the document, using the complete, unabbreviated form (e.g., full legal name, full title, full official designation). Do not shorten or paraphrase.
\end{enumerate}
}
\end{tcolorbox}

\begin{tcolorbox}[enhanced, breakable, boxrule=0.8pt, arc=4pt, fonttitle=\bfseries\small, fontupper=\small, before upper={\setlength{\parskip}{4pt}}, coltitle=white, colbacktitle=bluex, colback=bluex!3, colframe=bluex!70, left=4pt, right=4pt, top=4pt, bottom=4pt, title={Prompt for Formal Reasoning}]
{
You are a mathematical/physical reasoning assistant. Your task is to solve the math problem described in PROBLEM using the definitions and background provided in the context. All relevant definitions, lemmas, and background information are provided directly in the BACKGROUND section of the task. Refer to them as needed during your reasoning. Available tools: SymbolicReasoning (formal derivation), CodeExecutor (Python/SymPy computation), Compress (compress your derivation segment into a reusable summary), and Revise (undo to a previous step). You must perform at least one reasoning step (Think, SymbolicReasoning, or CodeExecutor) before calling Finish. Prefer SymbolicReasoning over Think for formal derivations.

WORKFLOW:
\vspace{-\parskip}
\begin{enumerate}[label=\arabic*., leftmargin=*, nosep, parsep=0pt]
    \item Use Think to analyse the problem and plan.
    \item Use SymbolicReasoning to carry out formal derivation steps (preferred for proofs and formal arguments).
    \item Use CodeExecutor to verify results with Python/ SymPy.
    \item Use Compress to compress your derivation into a summary.
    \item Use Revise to undo to a previous step if needed.
\end{enumerate}

Do NOT write markdown, headers, emojis, or any text outside the action call. Use LaTeX for mathematical expressions inside action parameters. Prefer SymbolicReasoning for formal derivations; Think is acceptable for brief analysis. At least one reasoning step before Finish. Use Compress to save intermediate results so that later sub-problems can reference them.
}
\end{tcolorbox}

\subsection{Baseline Evaluation Protocol}
For all RAG and memory baselines, we follow the evaluation protocol of MemoryArena~\citep{memoryarena2026}. Each method's storage $\mathcal{M}$ is initialized as empty at the start of each task. After each subtask finishes, the full interaction trace is written to $\mathcal{M}$ through the method's \texttt{Update} function. Before each action, we retrieve relevant entries from $\mathcal{M}$ using the current subtask trace as a query, i.e., previous actions and observations, and append the retrieved content to the agent context for action generation. We fix the retrieval TopK budget to 5 across all methods.

\subsection{Dataset Statistics}

\begin{table}[H]
\centering
\caption{Dataset statistics of MemoryArena.}
\label{tab:dataset_statistics}
\resizebox{\columnwidth}{!}{
\begin{tabular}{l ccc}
\toprule
\textbf{Domain} & \textbf{\#tasks} & \textbf{Avg. Trace Len.} & \textbf{Avg. Steps} \\
\midrule
Web Shopping & 150 & 24.53K & 98.56 \\
Travel Planning & 270 & 25.19K & 237.36 \\
Web Search & 221 & 101.92K & 81.98 \\
Math & 40 & 21.34K & 24.65 \\
Physics & 20 & 13.45K & 13.75  \\
\bottomrule
\end{tabular}
}
\end{table}

Table~\ref{tab:dataset_statistics} summarizes the statistics of the four MemoryArena~\citep{memoryarena2026} domains used in our evaluation. We report the number of tasks, the average trace length, and the average number of execution steps for each domain. Trace length denotes the token length of the full-task interaction trajectory, including actions and observations, while execution steps count the number of agent-environment interaction steps in this trajectory. Both statistics are measured from full-context rollouts. The results show that the tasks require agents to operate over long traces with many steps, making sustained state management essential for reliable performance.

\section{Analysis of Context Growth}
\label{app:context_growth}
\begin{figure}[H]
    \centering
    \includegraphics[width=1\linewidth]{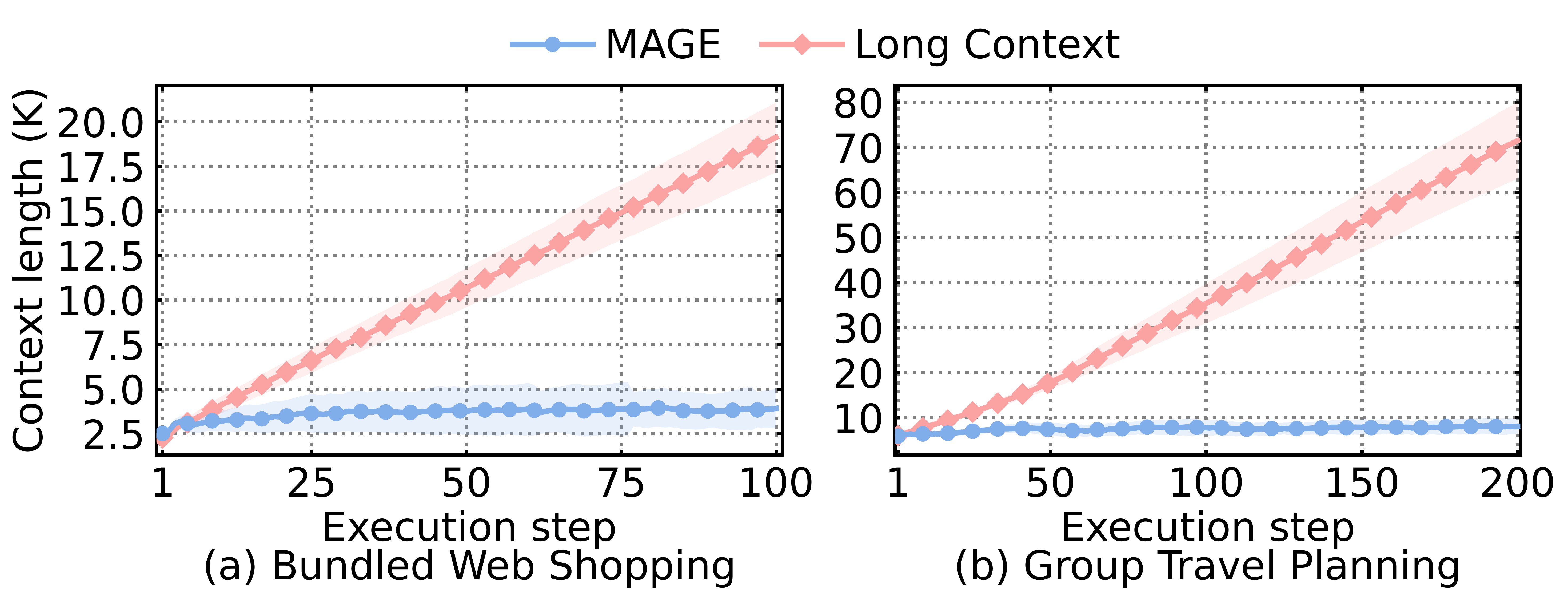}
    \caption{\textbf{Context Length vs. Execution Steps} (Average across tasks). \ourmethod bounds context growth through boundary-aware compression, whereas the full-context approach grows approximately linearly as execution steps accumulate.}
    \label{fig:context_growth}
\end{figure}

Figure~\ref{fig:context_growth} shows that the full-context approach grows approximately linearly with execution steps because it continuously appends the action-observation history at every step. In contrast, \ourmethod bounds context growth through boundary-aware \texttt{Compress}, which chunks completed bottom-layer segments (subgoals) into compact top-layer subgoal summaries while retaining only the recent unfinished trace. This reduction does not break execution-state integrity: as shown in Table~\ref{tab:main},~\ref{tab:qwen_moe} and~\ref{tab:gemma}, \ourmethod reduces token consumption while improving task success rate compared with the full-context approach, indicating that the boundary-aware compressed state still preserves the dependencies needed for downstream decisions.

\section{Results on DeepPlanning Benchmark}
\label{app:deepplanning}

MemoryArena~\citep{memoryarena2026} is a generic and representative benchmark for long-horizon interdependent tasks that require coherent execution-state management, spanning four domains with diverse action spaces, observation formats, and dependency structures. To further evaluate the generalizability of \ourmethod, we provide additional results on DeepPlanning~\citep{deepplanning2026}, a separate long-horizon planning benchmark comprising 240 tasks in total\footnote{We evaluate the English-version tasks, including 120 travel-planning tasks and 120 shopping-planning tasks.}, under the same experimental settings as the main experiments.

Unlike MemoryArena, DeepPlanning features a different action space based on structured API toolkits, less explicit subgoal structure without predefined task ordering, and complex global constraints, such as cross-item budget and schedule feasibility. As shown in Table~\ref{tab:deepplanning}, \ourmethod achieves the best performance on both DeepPlanning task categories while reducing token consumption compared with baselines. These results provide further evidence of the generalizability of \ourmethod across different long-horizon agentic tasks.

\begin{table}[t!]
\centering
\caption{\textbf{Results on DeepPlanning.} Best results are shown in \textbf{bold}, second best \underline{underlined}. \ourmethod achieves the best task accuracy while reducing the token consumption.}
\label{tab:deepplanning}
\resizebox{\linewidth}{!}{
\begin{tabular}{l cc cc}
\toprule
\multicolumn{1}{c}{\multirow{2}[2]{*}{\textbf{Method}}} & \multicolumn{2}{c}{\textbf{Travel Planning}} & \multicolumn{2}{c}{\textbf{Shopping Planning}} \\
\cmidrule(lr){2-3} \cmidrule(lr){4-5}
& Composite Score & \cellcolor{blue!5}\#tokens & Case Accuracy & \cellcolor{blue!5}\#tokens \\
\midrule
Full Context & 0.3248 & \cellcolor{blue!5}\textit{1129K} & 0.3944 & \cellcolor{blue!5}\textit{835K} \\
\midrule
HippoRAG2 & \underline{0.4295} & \cellcolor{blue!5}\textit{1056K} & 0.4417 & \cellcolor{blue!5}\textit{1002K} \\
MemoRAG & 0.4120 & \cellcolor{blue!5}\textit{995K} & 0.4400 & \cellcolor{blue!5}\textit{705K} \\
\midrule
Mem0 & 0.3666 & \cellcolor{blue!5}\textit{1292K} & 0.4333 & \cellcolor{blue!5}\textit{746K} \\
ReasoningBank & 0.3479 & \cellcolor{blue!5}\textit{720K} & 0.3718 & \cellcolor{blue!5}\textit{503K} \\
MemoryOS & 0.4090 & \cellcolor{blue!5}\textit{935K} & 0.3583 & \cellcolor{blue!5}\textit{710K} \\
SimpleMem & 0.4120 & \cellcolor{blue!5}\textit{1628K} & \underline{0.4833} & \cellcolor{blue!5}\textit{955K} \\
\midrule
\textbf{\ourmethod (Ours)} & \textbf{0.4818} & \cellcolor{blue!5}\textit{878K} & \textbf{0.4917} & \cellcolor{blue!5}\textit{524K} \\
\bottomrule
\end{tabular}
}
\end{table}

\section{Results on Additional Backend Models}
\label{app:backend_models}

To examine whether the benefits of \ourmethod generalize across different backend models, we evaluate Qwen3.6-35B-A3B~\citep{qwen3.6-35B-A3B} and Gemma4-31B-it~\citep{gemma4-31B} on MemoryArena~\citep{memoryarena2026}. As shown in Table~\ref{tab:qwen_moe} and Table~\ref{tab:gemma}, \ourmethod consistently improves SR over the baselines, achieving gains of $6.3$--$19.3$~pp on Qwen3.6-35B-A3B and $6.6$--$22.2$~pp on Gemma4-31B-it, respectively, averaged across tasks. It also reduces average token consumption by 45.0\% and 47.7\%, respectively, compared with the full-context approach. These results demonstrate that the benefits of \ourmethod generalize across backend models and are not tied to a particular model architecture.

\begin{table*}[t!]
\centering
\caption{\textbf{Results on MemoryArena (Qwen3.6-35B-A3B).} SR = Task Success Rate; PS = Task Progress Score; \#tokens = average token consumption per task. Best results in \textbf{bold}, second best \underline{underlined}. \ourmethod achieves the best task accuracy while reducing token consumption.}
\label{tab:qwen_moe}
\resizebox{\textwidth}{!}{
\setlength{\tabcolsep}{5pt}
\begin{tabular}{l ccc @{\hspace{14pt}} ccc @{\hspace{14pt}} ccc @{\hspace{14pt}} ccccc}
\toprule
& \multicolumn{3}{c@{\hspace{14pt}}}{\multirow{2}[2]{*}{\makecell{\textbf{Bundled}\\\textbf{Web Shopping}}}} & \multicolumn{3}{c@{\hspace{14pt}}}{\multirow{2}[2]{*}{\makecell{\textbf{Group}\\\textbf{Travel Planning}}}} & \multicolumn{3}{c@{\hspace{14pt}}}{\multirow{2}[2]{*}{\makecell{\textbf{Progressive}\\\textbf{Web Search}}}} & \multicolumn{5}{c}{\textbf{Formal Reasoning}} \\
\cmidrule(lr){11-15}
& & & & & & & & & & \multicolumn{2}{c}{Math} & \multicolumn{2}{c}{Physics} & ~ \\
\midrule
\textbf{Method} & SR & PS & \cellcolor{blue!5}\#tokens & SR & PS & \cellcolor{blue!5}\#tokens & SR & PS & \cellcolor{blue!5}\#tokens & SR & PS & SR & PS & \cellcolor{blue!5}\#tokens \\
\midrule
Full Context & \underline{0.2267} & \underline{0.6978} & \cellcolor{blue!5}\textit{4093K} & \underline{0.0440} & \underline{0.3890} & \cellcolor{blue!5}\textit{2755K} & 0.4163 & 0.2352 & \cellcolor{blue!5}\textit{6320K} & 0.2750 & 0.3277 & \textbf{0.6500} & \underline{0.6628} & \cellcolor{blue!5}\textit{1098K} \\
\midrule
HippoRAG2 & 0.1333 & 0.6189 & \cellcolor{blue!5}\textit{3537K} & 0.0122 & 0.3544 & \cellcolor{blue!5}\textit{3995K} & 0.3846 & 0.2547 & \cellcolor{blue!5}\textit{3823K} & \underline{0.3000} & 0.3616 & \underline{0.6000} & \underline{0.6628} & \cellcolor{blue!5}\textit{2438K} \\
MemoRAG & 0.1733 & 0.6622 & \cellcolor{blue!5}\textit{3232K} & 0.0037 & 0.3210 & \cellcolor{blue!5}\textit{3139K} & \underline{0.4208} & \underline{0.3016} & \cellcolor{blue!5}\textit{4159K} & \underline{0.3000} & 0.3277 & 0.5500 & 0.6047 & \cellcolor{blue!5}\textit{3148K} \\
\midrule
Mem0 & 0.1267 & 0.6089 & \cellcolor{blue!5}\textit{3852K} & 0.0037 & 0.2627 & \cellcolor{blue!5}\textit{2833K} & 0.3213 & 0.2169 & \cellcolor{blue!5}\textit{3050K} & \textbf{0.3250} & \textbf{0.3842} & \textbf{0.6500} & \textbf{0.6977} & \cellcolor{blue!5}\textit{1850K} \\
ReasoningBank & 0.0400 & 0.5089 & \cellcolor{blue!5}\textit{2891K} & 0.0068 & 0.2044 & \cellcolor{blue!5}\textit{1456K} & 0.1900 & 0.1450 & \cellcolor{blue!5}\textit{2036K} & 0.2750 & 0.3672 & 0.5000 & 0.6512 & \cellcolor{blue!5}\textit{1090K} \\
MemoryOS & 0.1200 & 0.6144 & \cellcolor{blue!5}\textit{3155K} & 0.0135 & 0.2540 & \cellcolor{blue!5}\textit{2697K} & 0.3122 & 0.2060 & \cellcolor{blue!5}\textit{3614K} & 0.2750 & \underline{0.3785} & 0.5500 & 0.6512 & \cellcolor{blue!5}\textit{1622K} \\
SimpleMem & 0.1333 & 0.6344 & \cellcolor{blue!5}\textit{6343K} & 0.0296 & 0.3344 & \cellcolor{blue!5}\textit{3947K} & 0.3891 & 0.2389 & \cellcolor{blue!5}\textit{5058K} & 0.2750 & 0.3503 & 0.5000 & \underline{0.6628} & \cellcolor{blue!5}\textit{2477K} \\
\midrule
\textbf{\ourmethod (Ours)} & \textbf{0.3067} & \textbf{0.7256} & \cellcolor{blue!5}\textit{2732K} & \textbf{0.0704} & \textbf{0.4708} & \cellcolor{blue!5}\textit{2045K} & \textbf{0.5204} & \textbf{0.4180} & \cellcolor{blue!5}\textit{2435K} & \textbf{0.3250} & 0.3559 & \textbf{0.6500} & \underline{0.6628} & \cellcolor{blue!5}\textit{853K} \\
\bottomrule
\end{tabular}
}
\end{table*}

\begin{table*}[t!]
\centering
\caption{\textbf{Results on MemoryArena (Gemma4-31B-it).} SR = Task Success Rate; PS = Task Progress Score; \#tokens = average token consumption per task. Best results in \textbf{bold}, second best \underline{underlined}. \ourmethod achieves the best task accuracy while reducing token consumption.}
\label{tab:gemma}
\resizebox{\textwidth}{!}{
\setlength{\tabcolsep}{5pt}
\begin{tabular}{l ccc @{\hspace{14pt}} ccc @{\hspace{14pt}} ccc @{\hspace{14pt}} ccccc}
\toprule
& \multicolumn{3}{c@{\hspace{14pt}}}{\multirow{2}[2]{*}{\makecell{\textbf{Bundled}\\\textbf{Web Shopping}}}} & \multicolumn{3}{c@{\hspace{14pt}}}{\multirow{2}[2]{*}{\makecell{\textbf{Group}\\\textbf{Travel Planning}}}} & \multicolumn{3}{c@{\hspace{14pt}}}{\multirow{2}[2]{*}{\makecell{\textbf{Progressive}\\\textbf{Web Search}}}} & \multicolumn{5}{c}{\textbf{Formal Reasoning}} \\
\cmidrule(lr){11-15}
& & & & & & & & & & \multicolumn{2}{c}{Math} & \multicolumn{2}{c}{Physics} & ~ \\
\midrule
\textbf{Method} & SR & PS & \cellcolor{blue!5}\#tokens & SR & PS & \cellcolor{blue!5}\#tokens & SR & PS & \cellcolor{blue!5}\#tokens & SR & PS & SR & PS & \cellcolor{blue!5}\#tokens \\
\midrule
Full Context & \underline{0.2800} & \underline{0.7178} & \cellcolor{blue!5}\textit{3103K} & \underline{0.1770} & 0.4756 & \cellcolor{blue!5}\textit{2295K} & 0.4027 & 0.2310 & \cellcolor{blue!5}\textit{5445K} & 0.4750 & \underline{0.4407} & \textbf{0.7500} & 0.7326 & \cellcolor{blue!5}\textit{1128K} \\
\midrule
HippoRAG2 & 0.1867 & 0.6744 & \cellcolor{blue!5}\textit{1920K} & 0.1185 & \underline{0.5297} & \cellcolor{blue!5}\textit{3655K} & 0.3982 & 0.2364 & \cellcolor{blue!5}\textit{3750K} & \underline{0.5000} & 0.4322 & \underline{0.7000} & 0.7093 & \cellcolor{blue!5}\textit{2264K} \\
MemoRAG & 0.1133 & 0.6133 & \cellcolor{blue!5}\textit{3244K} & 0.0424 & 0.3643 & \cellcolor{blue!5}\textit{3017K} & \underline{0.4118} & \underline{0.2608} & \cellcolor{blue!5}\textit{4459K} & \textbf{0.5500} & 0.4294 & \underline{0.7000} & \underline{0.7442} & \cellcolor{blue!5}\textit{3133K} \\
\midrule
Mem0 & 0.1200 & 0.6456 & \cellcolor{blue!5}\textit{2812K} & 0.0199 & 0.3406 & \cellcolor{blue!5}\textit{3116K} & 0.4027 & 0.2267 & \cellcolor{blue!5}\textit{3446K} & 0.4500 & 0.4266 & 0.6500 & 0.7093 & \cellcolor{blue!5}\textit{1886K} \\
ReasoningBank & 0.1200 & 0.5911 & \cellcolor{blue!5}\textit{1830K} & 0.0000 & 0.1953 & \cellcolor{blue!5}\textit{1466K} & 0.2443 & 0.1865 & \cellcolor{blue!5}\textit{2570K} & \underline{0.5000} & 0.4322 & 0.6000 & 0.7093 & \cellcolor{blue!5}\textit{1242K} \\
MemoryOS & 0.1200 & 0.6056 & \cellcolor{blue!5}\textit{1854K} & 0.0127 & 0.4004 & \cellcolor{blue!5}\textit{2746K} & 0.3575 & 0.2151 & \cellcolor{blue!5}\textit{3280K} & \textbf{0.5500} & 0.4266 & \underline{0.7000} & 0.7326 & \cellcolor{blue!5}\textit{1448K} \\
SimpleMem & 0.1933 & 0.6644 & \cellcolor{blue!5}\textit{3534K} & 0.0994 & 0.3632 & \cellcolor{blue!5}\textit{3365K} & 0.3710 & 0.2157 & \cellcolor{blue!5}\textit{5593K} & 0.4750 & 0.4237 & \underline{0.7000} & \textbf{0.7558} & \cellcolor{blue!5}\textit{2369K} \\
\midrule
\textbf{\ourmethod (Ours)} & \textbf{0.3467} & \textbf{0.7589} & \cellcolor{blue!5}\textit{1550K} & \textbf{0.2014} & \textbf{0.6429} & \cellcolor{blue!5}\textit{1518K} & \textbf{0.5339} & \textbf{0.3413} & \cellcolor{blue!5}\textit{2372K} & \underline{0.5000} & \textbf{0.4520} & \textbf{0.7500} & \underline{0.7442} & \cellcolor{blue!5}\textit{1105K} \\
\bottomrule
\end{tabular}
}
\end{table*}

\section{Case Studies}
\label{app:case_studies}
In this section, we present additional case studies to illustrate the effectiveness of \ourmethod in managing long-horizon tasks with interdependent steps, comparing it with baseline methods.

\begin{tcolorbox}[enhanced, breakable, boxrule=0.8pt, arc=4pt, fonttitle=\bfseries\small, fontupper=\small, before upper={\setlength{\parskip}{4pt}}, coltitle=white, colbacktitle=gray!65, colback=gray!4, colframe=gray!65, left=4pt, right=4pt, top=4pt, bottom=4pt, title={Case Study A: SimpleMem Retrieves Fragmented Memories Instead of Current State}]
\textbf{Task.} \texttt{task\_24\_subtask\_3}, Buy Product 4:

\textit{Compatibility notes:}
\vspace{-\parskip}
\begin{itemize}[leftmargin=*, nosep, parsep=0pt]
    \item Fudge pairs well with Chocolate.
    \item \textcolor{greenx}{Red pairs well with Gold}.
    \item Pink pairs well with Pearl.
    \item \textcolor{orangex}{Blue pairs well with Silver.}
    \item Yellow pairs well with Rainbow.
\end{itemize}

\textit{Avoid notes:}
\vspace{-\parskip}
\begin{itemize}[leftmargin=*, nosep, parsep=0pt]
    \item Fudge avoids Rainbow, Confetti, Pearl, Silver.
    \item \textcolor{greenx}{Red avoids Green, Silver, Rainbow}.
    \item Pink avoids Chocolate, Green, Orange.
    \item \textcolor{orangex}{Blue avoids Gold, Red, Orange.}
    \item Yellow avoids Silver, Gold.
\end{itemize}

\textit{Available options:}
\vspace{-\parskip}
\begin{itemize}[leftmargin=*, nosep, parsep=0pt]
    \item \textcolor{greenx}{A Gold Heart Sprinkles by Edible Party Supplies with 8 oz for baking and decorating cupcakes, cakes, cookies, and ice cream.}
    \item \textcolor{orangex}{A Mermaid Sprinkles by Cool Mom with silver, purple, and blue colors for baking and decorating cupcakes and cakes.}
    \item A Gold Sugar Crystals with 8 ounces by Sugar Deco.
    \item A Lieber's rainbow sprinkles with colorful jimmies for baking and decorating ice cream, 11 ounces.
    \item A Treasure Hunt Sprinkle Mix with colorful edible sprinkles for a birthday party in a 2oz size.
\end{itemize}

\textbf{Current state.} In the last subtask, SimpleMem and \ourmethod correctly purchased \textcolor{greenx}{AmeriColor Tulip Red AmeriMist Airbrush Food Color} (ASIN \texttt{B071S91WRZ}). Therefore, the correct transition for the current subtask is \textcolor{greenx}{Red $\rightarrow$ Gold}, and the ground-truth target is Gold Heart Sprinkles (ASIN \texttt{B08BFQ9B7T}) after comparing the price and rating between compatible options.
\end{tcolorbox}

\begin{tcolorbox}[enhanced, breakable, boxrule=0.6pt, arc=3pt, fonttitle=\bfseries\footnotesize, fontupper=\small, before upper={\setlength{\parskip}{4pt}}, coltitle=white, colbacktitle=orangex, colback=orangex!5, colframe=orangex!70, left=3pt, right=3pt, top=3pt, bottom=3pt, title={SimpleMem Trace}]
\textbf{Retrieved memories.}
\vspace{-\parskip}
\begin{itemize}[leftmargin=*, nosep, parsep=0pt]
    \item[-] [Memories 1] Content: Betty Crocker Rich \& Creamy Frosting - Cream Cheese - 16 oz - 3 pk with ... Topic: Final product purchase decision.
    \item[-] [Memories 2] Content: The third available cake base option is a \textcolor{orangex}{Duncan Hines Mermaid Cake Cup Vanilla Cake Mix with Sprinkles} weighing 2.15 Oz, identified by ASIN \texttt{B084JBT37C}, priced at \$6.99. ... Topic: Evaluation of Duncan Hines Mermaid Cake Cup.
    \item[-] [Memories 3] Content: The shopping task for selecting and purchasing a compatible frosting was successfully completed. Topic: Task completion status.
    \item[-] [Memories 4] Content: Strawberry cake toppers, lemon cupcake toppers, and lemon-flavored icing were excluded from consideration because ... Topic: Product option filtering.
    \item[-] [Memories 5] Content: A search for Betty Crocker ready-to-serve cream cheese frosting in an 8 pack returned ... Topic: Second product search results.
\end{itemize}

\textbf{Agent reasoning and actions.}

\vspace{-\parskip}
<think> ... Is there any chance the previous product is Red or Pink? \textcolor{orangex}{Memory 2 says "Duncan Hines Mermaid Cake Cup". Mermaid usually implies Blue/Silver/Green/Purple}. The compatibility note says ``\textcolor{orangex}{Blue pairs well with Silver}.'' This aligns perfectly. I will stick with this reasoning. ... </think> ... The previous product is likely the '\textcolor{orangex}{Duncan Hines Mermaid Cake Cup}' \textcolor{orangex}{(Blue)}, which pairs with Silver according to the rules. '\textcolor{orangex}{Mermaid Sprinkles}' is the only option explicitly containing Silver. ...

\textbf{Final purchase.} 

\vspace{-\parskip}
Mermaid Sprinkles by Cool Mom ... (\texttt{B07YTFYGXL}) at \$13.95 with rating 4.8.
\end{tcolorbox}

\begin{tcolorbox}[enhanced, breakable, boxrule=0.6pt, arc=3pt, fonttitle=\bfseries\footnotesize, fontupper=\small, before upper={\setlength{\parskip}{4pt}}, coltitle=white, colbacktitle=greenx, colback=greenx!5, colframe=greenx!70, left=3pt, right=3pt, top=3pt, bottom=3pt, title={\ourmethod Trace}]
\textbf{Compressed State.}
\vspace{-\parskip}
\begin{itemize}[leftmargin=*, nosep, parsep=0pt]
    \item[-] [Step 0] Product 1 Purchased: Gluten-Free Carrot Cake Mix. Price: \$14.99. ASIN: \texttt{B00T6NA7PA}. Rating: 4.5.
    \item[-] [Step 24] Product 2 Purchased: Betty Crocker Ready-to-Serve R \& C Frost Cream Cheese - 8 Pack. Price: \$34.27. ASIN: \texttt{B003GQP06O}. Rating: N.A..
    \item[-] [Step 31] Product 3 Purchased: \textcolor{greenx}{AmeriColor Tulip Red AmeriMist Airbrush Food Color}, 9 oz. Price: \$16.0. ASIN: \texttt{B071S91WRZ}. Rating: N.A..
\end{itemize}

\textbf{Agent reasoning and actions.}

\vspace{-\parskip}
... Prior purchase contains '\textcolor{greenx}{Red}'. According to rules, \textcolor{greenx}{Red pairs with Gold and avoids Green, Silver, Rainbow.} Option 1 (Gold Heart Sprinkles) contains Gold -> compatible. Option 2 (Mermaid Sprinkles) contains silver -> incompatible. Option 3 (Gold Sugar Crystals) contains Gold -> compatible. Option 4 (Lieber's rainbow sprinkles) contains rainbow -> incompatible. Option 5 (Treasure Hunt Sprinkle Mix) has no conflict keywords -> compatible. I will search all options to check prices, then select the highest-priced compatible one within the remaining budget of \$34.74. ...

\textbf{Final purchase.}

\vspace{-\parskip}
Gold Heart Sprinkles ... \texttt{B08BFQ9B7T} at \$14.95 with rating 4.4.
\end{tcolorbox}

\textbf{Case A analysis.} SimpleMem assembles the agent context from semantically similar entries, which causes state fragmentation: stale frosting memories and a non-final Mermaid candidate obscure the active execution state, namely that Product 3 was \textcolor{greenx}{AmeriColor Tulip Red}. This breaks execution-state integrity, so the agent maps Mermaid to \textcolor{orangex}{Blue/Silver/Green/Purple} and follows the wrong \textcolor{orangex}{Blue $\rightarrow$ Silver} transition rather than the correct \textcolor{greenx}{Red $\rightarrow$ Gold} transition. In contrast, \ourmethod reads the current execution state from the active path, preserving the purchase chain and selecting Gold Heart Sprinkles for the right reason.

\begin{tcolorbox}[enhanced, breakable, boxrule=0.8pt, arc=4pt, fonttitle=\bfseries\small, fontupper=\small, before upper={\setlength{\parskip}{4pt}}, coltitle=white, colbacktitle=gray!65, colback=gray!4, colframe=gray!65, left=4pt, right=4pt, top=4pt, bottom=4pt, title={Case Study B: Mem0 Retrieves Candidate Memories Instead of the Final Purchase}]
\textbf{Task.} \texttt{task\_128\_subtask\_1}, Buy Product 2, Select Matching Footrest:

\textit{Compatibility notes:}
\vspace{-\parskip}
\begin{itemize}[leftmargin=*, nosep, parsep=0pt]
    \item \textcolor{greenx}{Leather pairs well with Leather.}
    \item \textcolor{orangex}{Velvet pairs well with Velvet.}
\end{itemize}

\textit{Avoid notes:}
\vspace{-\parskip}
\begin{itemize}[leftmargin=*, nosep, parsep=0pt]
    \item \textcolor{greenx}{Leather avoids Velvet.}
    \item \textcolor{orangex}{Velvet avoids Leather.}
\end{itemize}

\textit{Available options:}
\vspace{-\parskip}
\begin{itemize}[leftmargin=*, nosep, parsep=0pt]
    \item \textcolor{greenx}{A blue PU leather ottoman storage chest with easy care for my living room.}
    \item A B FSOBEIIALEO storage ottoman cube with velvet material in beige, 2 pack.
    \item \textcolor{orangex}{A Best Master Furniture rectangle ottoman with grey velvet upholstery.}
    \item A Modway faux leather armchair and ottoman set in gray.
    \item A GARO leather storage ottoman bench with collapsible design in black.
\end{itemize}

\textbf{Current state.} In the previous subtask, Mem0 and \ourmethod correctly purchased \textcolor{greenx}{Blackjack Furniture Binion Leather Match Upholstered Modern Living Room Loveseat in Jet Black} (ASIN \texttt{B09H15V1KF}) at \$918.0 with rating 5. Therefore, the correct transition for the current subtask is \textcolor{greenx}{Leather $\rightarrow$ Leather}, and the ground-truth target is the blue PU leather Ottoman Storage Chest (ASIN \texttt{B09M76K4F8}) after comparing the prices among compatible leather options.
\end{tcolorbox}

\begin{tcolorbox}[enhanced, breakable, boxrule=0.6pt, arc=3pt, fonttitle=\bfseries\footnotesize, fontupper=\small, before upper={\setlength{\parskip}{4pt}}, coltitle=white, colbacktitle=orangex, colback=orangex!5, colframe=orangex!70, left=3pt, right=3pt, top=3pt, bottom=3pt, title={Mem0 Trace}]
\textbf{Retrieved memories.}
\vspace{-\parskip}
\begin{itemize}[leftmargin=*, nosep, parsep=0pt]
    \item[-] [Memories 1] Content: The 91-inch reversible sectional sofa (\texttt{B09PV5JQ91}) is available in multiple sizes and colors including greyish, beige, blue, grey, and light grey.
    \item[-] [Memories 2] Content: The second primary seating option the user is considering is a \textcolor{orangex}{modern 3-seater velvet couch with a tufted back in light gray}.
    \item[-] [Memories 3] Content: User verified the first seating option (\texttt{B09PV5JQ91}) is priced at \$889.00 with a 5-star rating and is now comparing it against the second option.
    \item[-] [Memories 4] Content: User is evaluating primary seating options with the goal of purchasing the highest-priced item, using ratings as a tiebreaker for compatible choices.
    \item[-] [Memories 5] Content: User searched for a \textcolor{orangex}{modern 3-seater velvet couch} with tufted back in light gray and found matching product \texttt{B09H2XT45M}, priced at \$599.99.
\end{itemize}

\textbf{Agent reasoning and actions.}

\vspace{-\parskip}
... Based on the retrieved memories, the user's primary seating selection involves \textcolor{orangex}{velvet furniture} (modern 3-seater velvet couch \texttt{B09H2XT45M} and 91-inch reversible sectional sofa in velvet grey \texttt{B09PV5JQ91}). According to compatibility notes, \textcolor{orangex}{Velvet pairs well with Velvet, and Velvet avoids Leather}. Therefore, I need to select a \textcolor{orangex}{velvet footrest}. ... The other options (PU leather, faux leather, leather) are not compatible with velvet. ...

\textbf{Final purchase.}

\vspace{-\parskip}
Best Master Furniture SH003 Rectangle Ottoman, Grey Velvet ... (\texttt{B08QYLGB9R}) at \$362.92.
\end{tcolorbox}

\begin{tcolorbox}[enhanced, breakable, boxrule=0.6pt, arc=3pt, fonttitle=\bfseries\footnotesize, fontupper=\small, before upper={\setlength{\parskip}{4pt}}, coltitle=white, colbacktitle=greenx, colback=greenx!5, colframe=greenx!70, left=3pt, right=3pt, top=3pt, bottom=3pt, title={\ourmethod Trace}]
\textbf{Compressed State.}
\vspace{-\parskip}
\begin{itemize}[leftmargin=*, nosep, parsep=0pt]
    \item[-] [Step 0] Product 1 Purchased: \textcolor{greenx}{Blackjack Furniture Binion Leather Match Upholstered Modern Living Room Loveseat, Jet Black}. Price: \$918.0. ASIN: \texttt{B09H15V1KF}. Rating: 5.
\end{itemize}

\textbf{Agent reasoning and actions.}

\vspace{-\parskip}
... Product 1 contains \textcolor{greenx}{``Leather''}. Compatibility rules state \textcolor{greenx}{Leather pairs well with Leather and Leather avoids Velvet}. I will filter options to those containing leather and not velvet. Options 1, 4, and 5 match this criterion. ... Option 1 (PU leather): compatible, price \$343.89. Option 2 (velvet): incompatible. Option 3 (velvet): incompatible. Option 4 (faux leather): compatible, price \$271.34. Option 5 (leather): compatible, price \$42.88. Among compatible options, Option 1 has the highest price. ...

\textbf{Final purchase.}

\vspace{-\parskip}
Ottoman Storage Chest Ottomans for Doorway, PU Leather Easy Care \& Easy Clean Footrest (Blue) ... \texttt{B09M76K4F8} at \$343.89.
\end{tcolorbox}

\textbf{Case B analysis.} Mem0 retrieves semantically related memories about primary-seating candidates rather than the final purchased state. The retrieved context contains multiple \textcolor{orangex}{velvet}-related memories, but omits the active purchase \textcolor{greenx}{Blackjack Leather Loveseat} (ASIN \texttt{B09H15V1KF}). As a result, the agent reconstructs the previous state as \textcolor{orangex}{Velvet} and follows the wrong \textcolor{orangex}{Velvet $\rightarrow$ Velvet} transition, selecting a grey velvet ottoman and excluding the leather options. In contrast, \ourmethod reads the current execution state from the active path, preserves the fact that Product 1 was \textcolor{greenx}{Leather}, and follows \textcolor{greenx}{Leather $\rightarrow$ Leather} to buy the correct PU leather ottoman.

\end{document}